\documentclass[11pt]{article}
\usepackage{coling2018}
\usepackage{times}
\usepackage{url}
\usepackage{latexsym}

\usepackage{graphicx}
\usepackage{comment}
\usepackage{paralist}
\usepackage{footmisc}
\usepackage{wrapfig}
\usepackage{enumitem}

\usepackage{multirow}
\usepackage{xspace} 
\usepackage{fancyvrb} 
\usepackage{caption} 
\usepackage[color=red!5]{todonotes}
\let\todonote\todo
\usepackage[skins]{tcolorbox}
\usepackage{xstring}
\usepackage{soul}
\usepackage[normalem]{ulem} 
\usepackage{algpseudocode,algorithm} 
\usepackage{tikz-qtree-compat}
\usetikzlibrary{shapes, fit, arrows.meta}
\usepackage{xcolor,colortbl} 
\usepackage{tabularx}
\usepackage{flafter} 

\usepackage{adjustbox}

\algnewcommand{\LineComment}[1]{\State \(\triangleright\) #1} 

\AtBeginEnvironment{algorithmic}{\scriptsize} 
\algrenewcommand\alglinenumber[1]{\scriptsize #1:} 

\tikzset{>=latex}

\setlist[description]{%
  labelindent=0em, 
  leftmargin=1em,
}

\renewcommand{\todo}[1]{\todonote[size=\tiny]{#1}{\textcolor{red}{(TODO: #1)}}}
\newcommand{\annot}[2][]{\todonote[size=\tiny]{#1}{\textcolor{red}{(#2)}}}

\newcommand{\texttuple}[3]{$\langle$\textit{#1}; \textit{#2}; \textit{#3}$\rangle$}

\newsavebox{\fmbox}

\title{Graphene: A Context-Preserving Open Information Extraction System}

\author{Matthias Cetto\textsuperscript{1}, Christina Niklaus\textsuperscript{1}, Andr\'{e} Freitas\textsuperscript{2}, \and Siegfried Handschuh\textsuperscript{1} \\
  \textsuperscript{1} Faculty of Computer Science and Mathematics, University of Passau\\
  {\tt \{matthias.cetto, christina.niklaus, siegfried.handschuh\}{\tt @uni-passau.de}}\\
  \textsuperscript{2} School of Computer Science, University of Manchester\\
  {\tt andre.freitas@manchester.ac.uk}
\\}

\date{}

\begin{document}
\maketitle
\begin{abstract}
We introduce Graphene, an Open IE system whose goal is to generate accurate, meaningful and complete propositions that may facilitate a variety of downstream semantic applications. For this purpose, we transform syntactically complex input sentences into clean, compact structures in the form of core facts and accompanying contexts, while identifying the rhetorical relations that hold between them in order to maintain their semantic relationship. In that way, we preserve the context of the relational tuples extracted from a source sentence, generating a novel lightweight semantic representation for Open IE that enhances the expressiveness of the extracted propositions.
\end{abstract}

\section{Introduction}
\label{intro}

%
%
\blfootnote{
    %
    %
    %
    %
     \hspace{-0.65cm}  
     This work is licenced under a Creative Commons 
     Attribution 4.0 International Licence.
     Licence details:
     \url{http://creativecommons.org/licenses/by/4.0/}
    %
    %
}

Information Extraction (IE) is the task of turning the unstructured information expressed in natural language (NL) text into a structured representation in the form of relational tuples consisting of a set of arguments and a phrase denoting a semantic relation between them: \texttuple{arg1}{rel}{arg2}. Unlike traditional IE methods, Open IE is not limited to a small set of target relations known in advance, but rather extracts all types of relations found in a text. In that way, it facilitates the domain-independent discovery of relations extracted from text and scales to large, heterogeneous corpora such as the Web. Since its introduction by \newcite{Banko07}, a large body of work on the task of Open IE has been described. By analyzing the output of state-of-the-art systems (e.g., \cite{Mausam12,DelCorro13,Angeli15}), we observed three common shortcomings. 

First, relations often span over long nested structures or are presented in a non-canonical form that cannot be easily captured by a small set of extraction patterns. Therefore, such relations are commonly missed by state-of-the-art approaches. Second, current Open IE systems tend to extract propositions with long argument phrases that can be further decomposed into meaningful propositions, with each of them representing a separate fact. Overly specific constituents that mix multiple - potentially semantically unrelated - propositions are difficult to handle for downstream applications, such as question answering (QA) or textual entailment tasks. Instead, such approaches benefit from extractions that are as compact as possible. Third, state-of-the-art Open IE systems lack the expressiveness needed to properly represent complex assertions, resulting in incomplete, uninformative or incoherent propositions that have no meaningful interpretation or miss critical information asserted in the input sentence.

To overcome these limitations, we developed an Open IE framework called "Graphene" that transforms syntactically complex NL sentences into clean, compact structures that present a canonical form which facilitates the extraction of accurate, meaningful and complete propositions. The contributions of our work are two-fold. First, to remove the complexity of determining intricate predicate-argument structures with variable arity from syntactically complex input sentences, we propose a two-layered transformation process consisting of a clausal and phrasal disembedding layer. It removes clauses and phrases that convey no central information from the input and converts them into independent context sentences, thereby reducing the source sentence to its main information. In that way, the input is transformed into a \textbf{novel hierarchical representation in the form of core facts and accompanying contexts}. Second, we \textbf{identify the rhetorical relations by which core sentences and their associated contexts are connected in order to preserve their semantic relationship}. These two innovations enable us to enrich extracted relational tuples of the form \texttuple{arg1}{rel}{arg2} with contextual information that further specifies the tuple and to establish semantic links between them, resulting in a novel lightweight semantic representation for Open IE that provides highly informative extractions and thus supports their interpretability in downstream applications. The source code is available at \url{https://github.com/Lambda-3/Graphene}.



\section{The System in a Nutshell}



\begin{figure*}[ht]
\centering

\begin{minipage}{0.17\textwidth}
\centering
Input-Document\\
\vspace{0.3cm}
\begin{tikzpicture}[scale=0.55, every node/.style={align=center, transform shape}]
\node[rectangle, solid, draw=black, text width=0.85*\columnwidth, rounded corners=5pt]{
[...] Although the Treasury will announce details of the November refunding on Monday, the funding will be delayed if Congress and President Bush fail to increase the Treasury's borrowing capacity. [...]
};
\end{tikzpicture}
\end{minipage}%
\hspace{-0.5cm}
$\rightarrow$
\begin{minipage}{0.35\textwidth}
\centering
Transformation Stage\\
\vspace{0.3cm}
\begin{tikzpicture}[scale=0.55, level distance=2cm, sibling distance=0cm, every tree node/.style={align=center, transform shape}]
\Tree [
.\node[style={draw,rectangle}] {DOCUMENT-ROOT}; 
  \edge node[midway, right] {core}; [
      .\node [style={draw,rectangle}] {Coordination\\\textit{Contrast}};
            \edge node[midway, left] {core}; [.\node(a)[label=below:TEMPORAL(on Monday)]{The Treasury will\\announce details of\\the November refunding.};]
            \edge node[midway, right] {core}; [
              .\node [style={draw,rectangle}] {Subordination\\\textit{Condition}};
                \edge node[midway, left] {core}; [.\node(b){The funding\\will be delayed.};]
                \edge node[midway, right] {context}; [
                  .\node [style={draw,rectangle}] {Coordination\\\textit{List}};
                    \edge node[midway, left] {core}; [.\node(c){Congress fail to\\increase the Treasury's\\borrowing capacity.};]
                    \edge node[midway, right] {core}; [.\node(d){President Bush fail to\\increase the Treasury's\\borrowing capacity.};]
                ]
            ]
        ]
    ]
]
\end{tikzpicture}
\end{minipage}%
$\rightarrow$
\hspace{0.5cm}
\begin{minipage}{0.35\textwidth}
\centering
Relation Extraction\\
\vspace{0.3cm}
\begin{tikzpicture}[scale=0.55, every node/.style={align=center, transform shape}]
    \node(A)[text width=0.6*\linewidth, label=below:TEMPORAL(on Monday)]{The Treasury will announce details [...]};
    \node(B)[below of=A, yshift=-1cm, text width=0.6*\linewidth]{The funding will be delayed.};
    \node(C)[below of=B, yshift=-0.5cm, text width=0.6*\linewidth]{Congress fail to increase [...]};
    \node(D)[below of=C, yshift=-0.5cm, text width=0.6*\linewidth]{President Bush fail to increase [...]};

    \node(a)[right=1 of A, text width=0.6*\linewidth,  label=below:TEMPORAL(on Monday)]{\texttuple{The Treasury}{will announce}{details [...]}};
    \node(b)[right=1 of B, text width=0.6*\linewidth]{\texttuple{The funding}{will be delayed}{}};
    \node(c)[right=1 of C, text width=0.6*\linewidth]{\texttuple{Congress}{fail}{to increase [...]}};
    \node(d)[right=1 of D, text width=0.6*\linewidth]{\texttuple{President Bush}{fail}{to increase [...]}};

    \draw[solid, <->] (A)..controls +(west:3) and +(west:3)..([yshift=10]B) node [left, midway] () {Contrast};
    \draw[solid, ->] (B)..controls +(west:3) and +(west:3)..([yshift=5]C) node [left, midway] () {Condition};
    \draw[solid, ->] ([yshift=-10]B)..controls +(west:3) and +(west:3)..([yshift=-5]D) node [left, midway] () {Condition};
    \draw[solid, <->] ([yshift=-5]C)..controls +(west:2.5) and +(west:2.5)..([yshift=5]D) node [right, midway] () {List};

    \draw[solid, ->] (A) to (a);
    \draw[solid, ->] (B) to (b);
    \draw[solid, ->] (C) to (c);
    \draw[solid, ->] (D) to (d);

\end{tikzpicture}\\
\end{minipage}

\caption{Extraction workflow for an example sentence.}
\label{fig:workflow}
\end{figure*}
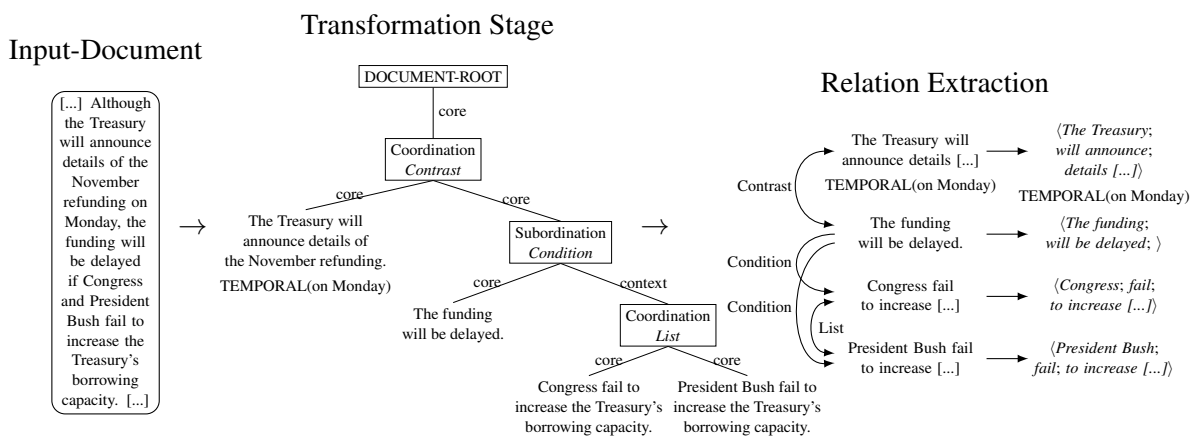

Graphene makes use of a two-layered transformation stage consisting of a clausal and phrasal disembedding layer, which is followed by a final relation extraction (RE) stage. It takes a text document as an input and returns a set of semantically typed and interconnected relational tuples. The workflow of our approach is displayed in Figure~\ref{fig:workflow}.

\subsection{Transformation Stage}
During the transformation process, source sentences that present a complex linguistic structure are converted into a hierarchical representation of core facts and associated contexts that are connected by rhetorical relations capturing their semantic relationship similar to Rhetorical Structure Theory (RST) \cite{mann1988rhetorical}. These compact, syntactically sound structures ease the problem of recognizing predicate-argument relations that are contained in the input without losing their semantic dependencies. 

\paragraph{Clausal Disembedding.}
In the clausal disembedding layer, we split up complex multi-clause sentences that are composed of \textit{coordinated} and \textit{subordinated clauses}, \textit{relative clauses}, or \textit{attributions} into simpler, stand-alone sentences that contain one clause each. This is done in a recursive fashion so that we obtain a hierarchical structure of the transformation process comparable to the diagrams used in RST. As opposed to RST, however, the transformation process is carried out in a top-down fashion, starting with the input document and using a set of hand-crafted syntactic rule patterns that define how to split up, transform and recurse on complex syntactic patterns\footnote{The complete rule set can be found online: \url{https://github.com/Lambda-3/Graphene/blob/master/wiki/supplementary/syntactic-simplification-patterns.pdf}}. Each split will create two or more simplified sentences that are connected with information about (1) their constituency type depicting their semantic relevance (\textit{coordinate} or \textit{subordinate}) and (2) the rhetorical relation that holds between them. The constituency type infers the concept of nuclearity from RST, where coordinate sentences (which we call \textit{core sentences}) represent nucleus spans that embody the central part of information, while subordinate sentences (\textit{context sentences}) represent satellite spans that provide background information on the nucleus. 
The classification of the rhetorical relations is based on both syntactic and lexical features. While former are manifested in the phrasal composition of a sentence's phrasal parse tree, latter rely on a set of manually defined cue phrases. In this way, a hierarchical tree representation of the recursive transformation process for the whole document is constructed which we denote as \textit{discourse tree}. Its leaf nodes represent the simplified sentences that were generated during the clausal disembedding layer.

\paragraph{Phrasal Disembedding.}

After recursively dividing multi-clause sentences into stand-alone sentences that contain one clause each, they are further simplified on a phrasal level. For this purpose, sentences are processed separately and transformed into simpler structures by extracting the following phrasal components from the input: \textit{prepositional phrases, participial phrases, adjectival/adverbial phrases, appositive phrases, lead noun phrases, coordinations of verb phrases, enumerations of noun phrases} and \textit{purposes}. This task is assisted by the sentence simplification system described in \newcite{Niklaus2016}.

\subsection{Relation Extraction}
After the transformation stage, RE is performed by using the simplified sentences as an input. The framework is designed to accept any type of RE implementation which is able to extract relational tuples from single sentences. The identified rhetorical relations from the transformation stage are then mapped to the corresponding relational tuples in the form of simple and linked contextual arguments (see Section~\ref{sec:outputFormat}). As a result, different approaches for RE can be complemented with contextual information that further specifies the extracted relational tuples. In that way, a new layer of semantics is added to the task of RE that can be used in other NLP tasks (see Section~\ref{sec:othertasks}).

\section{Output Format}
\label{sec:outputFormat}

In order to represent contextual relations between propositions, the default representation of a relational tuple of the form \texttuple{arg1}{rel}{arg2} needs to be extended. Therefore, we present a novel lightweight semantic representation for Open IE that is both machine processable and human readable.
It extends a binary subject-predicate-object tuple $t \gets (rel, arg_{subj}, arg_{obj})$ with: a unique identifier $id$; information about the contextual hierarchy, the so-called \textit{context-layer} $cl$; and two sets of semantically classified contextual arguments $C_S$ (\textit{simple contextual arguments}) and $C_L$ (\textit{linked contextual arguments}), yielding the final representation of $(id, cl, t, C_S, C_L)$ tuples. The \textit{context-layer} $cl$ encodes the contextual hierarchy of core and contextual facts. Propositions with a context-layer of $0$ carry the core information of a sentence, whereas propositions with a context-layer of $cl > 0$ provide contextual information about propositions with a context-layer of $cl - 1$. Both types of contextual arguments $C_S$ and $C_L$ provide (semantically classified) contextual information about the statement expressed in $t$. Whereas a simple contextual argument $c_S \in C_S, c_S \gets (s, r)$ contains a textual expression $s$ that is classified by the semantic relation $r$, a linked contextual argument $c_L \in C_L, c_L \gets (id(z), r)$ refers to the content expressed in another proposition $z$.

To facilitate the inspection of the extracted propositions, a human-readable format, called \textit{RDF-NL}, is generated by Graphene (see Figure~\ref{fig:GrapheneOutput}). In this format, propositions are grouped by sentences in which they occur and are represented by tab-separated strings for the identifier $id$, context-layer $cl$ and the core extraction that is represented by the binary relational tuple $t \gets (rel, arg_{subj}, arg_{obj})$: subject argument $arg_{subj}$, relation name $r$ and object argument $arg_{obj}$. 
Contextual arguments ($C_S$ and $C_L$) are indicated by an extra indentation level to their parent tuples. The representation of a contextual argument consists of a type string and a tab-separated content. The type string encodes both the context type (\texttt{S} for a simple contextual argument $c_S \in C_S$ and \texttt{L} for a linked contextual argument $c_L \in C_L$) and the classified semantic relation (e.g. \textit{Cause}, \textit{Purpose}), if present. The content of a simple contextual argument is the textual expression, whereas the content of a linked contextual argument is the identifier of the target proposition.

\begin{figure}[htbp]
	\centering
    \begin{adjustbox}{max width=\linewidth}
    \begin{BVerbatim}[fontsize=\scriptsize]      
Although the Treasury will announce details of the November refunding on Monday, the funding
will be delayed if Congress and President Bush fail to increase the Treasury's borrowing capacity.

#1    0    the Treasury    will announce    details of the November refunding
    S:TEMPORAL    on Monday
    L:CONTRAST     #2

#2    0    the funding    will be delayed 
    L:CONTRAST     #1
    L:CONDITION    #3
    L:CONDITION    #4

#3    1    Congress    fail    to increase the Treasury 's borrowing capacity

#4    1    president Bush    fail    to increase the Treasury 's borrowing capacity

    \end{BVerbatim}
    \end{adjustbox}
    \caption{Proposed representation format (RDF-NL) - human readable representation.}
	\label{fig:GrapheneOutput}
\end{figure}

Besides, the framework can materialize its relations into a graph serialized under the N-Triples{\footnote{\url{https://www.w3.org/TR/n-triples}}} specification of the Resource Description Framework (RDF) standard. In that way, the consumption of the extracted relations by downstream applications is facilitated.
A detailed description as well as some examples of the machine-readable RDF format are available online\footnote{\url{https://github.com/Lambda-3/Graphene/blob/master/wiki/RDF-Format.md}}.

\section{Usage}
Graphene can be either used as a Java API, imported as a Maven dependency, or as a service which we provide through a command line interface or a REST-like web service that can be deployed via docker. A demonstration video is available online\footnote{\url{https://asciinema.org/a/bvhgIP8ZEgDwtmRPFctHyxALu?speed=3}}.

\section{Benchmarking}

\begin{wrapfigure}{r}{7.5cm}
	\centering
	\includegraphics[width=7.5cm]{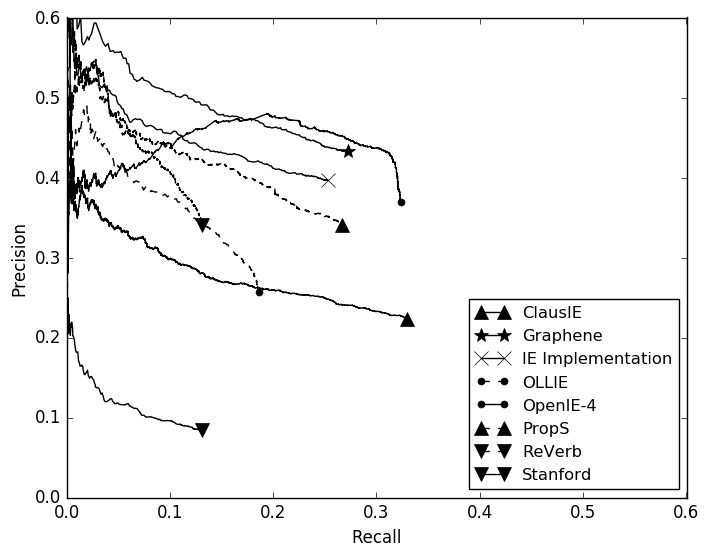}  
  \caption{Performance of Graphene.}
  \label{fig:OIEBenchmark}
\end{wrapfigure}
We evaluated the performance of our Open IE system Graphene using the benchmark framework proposed in \newcite{Stanovsky2016EMNLP}, which is based on a QA-Semantic Role Labeling corpus with more than 10,000 extractions over 3,200 sentences from Wikipedia and the Wall Street Journal\footnote{available under \url{https://github.com/gabrielStanovsky/oie-benchmark}}. This benchmark allowed us to compare our framework with a set of state-of-the-art Open IE approaches in recall and precision (see Figure~\ref{fig:OIEBenchmark}). With a score of 50.1\% in average precision, Graphene achieves the best performance of all the systems in extracting accurate tuples. Considering recall, our framework (27.2\%) is able to compete with the best-performing baseline approaches (32.5\% and 33.0\%). The interested reader can refer to \newcite{Cetto2018} for more details.

\section{Application Scenarios of the Lightweight Semantic Open IE representation}
\label{sec:othertasks}
The resulting lightweight semantic representation of the source text in the form of a two-layered hierarchy of semantically-linked relational tuples can be used to facilitate a variety of artificial intelligence tasks, such as building QA systems, creating text summarization applications or supporting semantic inferences.

For example, QA systems could build upon the semantically typed and interconnected relational tuples produced by our Open IE system Graphene to investigate the dependencies between extracted propositions (such as causalities, attributions and local or temporal contexts) and map specific question types to the corresponding semantic relationships when querying the underlying data. Based on the example given in Figure~\ref{fig:GrapheneOutput}, one can imagine the following user query: 

\begin{center}
\textit{\uline{Under which circumstances} will the funding be delayed?}
\end{center}

Here, the system could infer from the interrogative expression \textit{"Under which circumstances?"} to search for propositions that are linked to the extraction stating that \texttuple{the funding}{will be delayed}{$\emptyset$} by a conditional (\texttt{CONDITION}) relation. Accordingly, in this scenario the system is expected to return propositions \#3 and \#4 of Figure~\ref{fig:GrapheneOutput}. 

\section{Conclusion}
We presented Graphene, an Open IE system that transforms sentences which present a complex linguistic structure into a novel hierarchical representation in the form of core facts and accompanying contexts which are connected by rhetorical relations capturing their semantic relationship. In that way, the input text is turned into clean, compact structures that show a canonical form, thus facilitating the extraction of accurate, meaningful and complete propositions based on a novel lightweight semantic representation consisting of a set of semantically typed and interconnected relational tuples. In the future, we aim to port this idea to languages other than English.


\bibliographystyle{acl}
\bibliography{coling2018}

\end{document}